\documentclass[12pt]{article}
\usepackage{amsmath,amssymb,bbm,dsfont,mathrsfs}
\usepackage{fullpage,graphicx}

\newtheorem{theorem}{Theorem}

\newtheorem{lemma}{Lemma}

\newcommand{\startmat}[1]{\left[\begin{array}{#1}}
\newcommand{\closemat}{\end{array}\right]}

\newcommand{\be}{\begin{equation}}
\newcommand{\ee}{\end{equation}}
\newcommand{\I}{{\cal I}}
\newcommand{\N}{{\cal N}}
\newcommand{\M}{{\cal M}}
\newcommand{\B}{{\cal B}}

\newcommand{\tl}{{\triangleleft}}
\begin{document}
\title{Distributed parameter estimation of discrete hierarchical models via marginal likelihoods }
\author{\textsc{By H\'{e}l\`{e}ne Massam\footnote{H. Massam gratefully acknowledges support from NSERC Discovery Grant No A8947.} and Nanwei Wang}\\{\it Department of Mathematics and Statistics, York University,}\\ {\it Toronto, ON M3J 1P3, Canada}}

\date{October 17, 2013}

\maketitle
{\abstract We consider discrete graphical models Markov with respect to a graph $G$ and propose two distributed marginal methods to estimate the maximum likelihood estimate of the canonical parameter of the model. Both methods are based on a relaxation of the marginal likelihood obtained by considering the density of the variables represented by a vertex $v$ of $G$  and a neighborhood. The  two methods differ by the size of the neighborhood of $v$. We show that the estimates are consistent and that those obtained with the larger neighborhood have  smaller asymptotic variance than the ones obtained through the smaller neighborhood.
}
\vspace{4mm}

\vspace{.4cm}


\section{Introduction}
Maximum likelihood estimation of the parameter is an intrinsic part of learning in Markov random fields and much research has been devoted to finding efficient algorithms to obtain a reasonably  accurate estimate of the parameter. In this paper, we will consider  the class of discrete graphical models embedded in the larger class of discrete hierarchical models. 
A number $N$ of individuals are classified according to criteria indexed by a finite set $V$. Each variable $X_v, v\in V$  takes values in a finite set $\I_v$. The   data  is gathered in a contingency table and we assume that the cell counts follow a multinomial distribution Markov with respect to a given undirected graph $G$. The distribution of the cell counts belongs to a natural exponential family with density of the general form
\begin{eqnarray*}
f(t; \theta)=\exp \{\langle \theta, t \rangle-Nk(\theta)\}
\end{eqnarray*}
where $\langle \theta, t \rangle$ denotes the inner product of the variable $t(x)$ and the canonical parameter $\theta=(\theta_j, j\in J)$ in ${\mathbb{R}}^{|J|}$ where $J$ is a subset of the set of cells which characterizes the given model. 
The log partition function or cumulant generating function $k(\theta)$  is usually intractable when the number of variables is large. 
  Therefore, even though the likelihood function is a convex function of the parameter, the traditional convex optimization methods using the derivative of the likelihood cannot be used. Recent computationally efficient approximate iterative methods usually involve  loopy belief propagation. A thorough review of the literature for discrete models as well as a new approach called Constrained Approximate Maximum Entropy Learning can be found in  Ganapathi et al. (2012).   
  \vspace{3mm}
  
   A radically different  type of maximum likelihood estimation method  for graphical models has been proposed in the last two years. It involves distributive learning. The essence of these methods is to  estimate subsets of the components of $\theta$ via local  marginal or conditional likelihoods where "local" means relative to one variable $X_v$. For Gaussian graphical models, Wiesel and Hero (2012) gave two methods. One of these methods is based on the pseudo-likelihood (or composite likelihood) and consists in maximizing the conditional likelihood of $X_v$ given $X_{\N_v}$ where $\N_v$ denotes the set of neighbours of $v$ in the graph $G$ underlying the model. The other method consists in maximizing a likelihood which is a relaxation of the marginal likelihood of $X_{v\cup \N_v}$. In either method, a given component of the parameter $\theta$ may have estimates coming from several $v\in V$. These estimates are either combined  through some convex combination or they are improved upon through an iterative algorithm, the Alternative Direction Method of Multipliers (ADMM, see Boyd et al. 2010).  Meng, Wei, Wiesel and Hero (2013) refined  the relaxed marginal method by including in their local marginal model,  the variables indexed not only by the vertex $v$ and its neighbours but also by the neighbours of the neighbours. They observed that with this new ''two-hop"  method, the estimates of $\theta$ were  very accurate and no improvement involving  ADMM or other adjustments was necessary. 
   For discrete graphical loglinear models, Qiang and Liu (2012) gave the equivalent of the pseudo-likelihood  method of Wiesel and Hero (2012) together  with a fine asymptotic analysis of the precision of the estimates. In that paper, the local pseudo-likelihood estimates are improved using either consensus or ADMM.
   \vspace{3mm}

   In this paper,  we develop the equivalent of the marginal one-hop (involving a vertex $v$ and its immediate neighbours)  method of Wiesel and Hero (2012) and  the marginal two-hop (involving  $v$, its neighbours and the neighbours of the neighbours)  method of  Meng, Wei, Wiesel and Hero (2013) for the  class of  discrete graphical log-linear models. We will see that in the marginalization process, one moves from the class of graphical models to the larger class of hierarchical models.  We show that the  one-hop and  two-hop estimates  are asymptotically consistent and that the asymptotic variance of the two-hop estimate is always smaller than that of the one-hop estimate and this, of course, implies higher accuracy. Through numerical experiments, we evaluate the performance of these new estimates. We find that the one-hop marginal estimate is extremely fast and as accurate as the pseudo-likelihood estimate. The two-hop method is somewhat slower than the pseudo-likelihood estimate but is extremely accurate and does not need further consensus or ADMM adjustments. The sample variances of the global, one-hop marginal, pseudo-likelihood and two-hop marginal estimates are graphed versus  sample size,  for models Markov with respect to a four-neighbour $k\times k$ lattices, $k=4$ and $10$ and a random graph with 100 vertices. The relationship between the asymptotic variance of the various estimates is reflected in the graph of the sample variances.
 \vspace{3mm}

Before proceeding to describe our methods, we note that our results are given under the assumption that the maximum likelihood estimate (henceforth abbreviated mle) of $\theta$  exists, that is, the mle of $\theta$ is such that the corresponding estimates of cell probabilities are positive. This is an important remark whatever the estimation method might be. Indeed, when the graph is large and the data sparse, the mle might not exist. 
The existence of the mle in the loglinear model has been studied in detail in Fienberg et al.(2012).  
 \vspace{3mm}

The paper is organized as follows. In the next section, we formulate the problem and establish our notation. In Section 3, we derive the two-hop estimate and its asymptotic properties. Numerical results are given in the last section.
\section{Preliminaries}
\subsection{ Discrete graphical and hierarchical loglinear models}
Let $V$ be a finite set of indices representing $p=|V|$ criteria. Let $X=(X_v,\;v\in V)$ be a multivariate discrete random variable such that each variable $X_v$  takes its  values in a finite set $I_v$.
If $N$ individuals are classified according to these $|V|$ criteria, the resulting counts are gathered in a contingency table such that
$$I=\prod_{v\in V}I_{v}$$
is the  set of cells $i=(i_v,\;v\in V)$.  For $D\subset V$, $i_D$ denotes the marginal cell $i_D=(i_v, v\in D)$.

Let $\mathcal{D}$ be a family of non empty subsets of $V$ such that $D\in \mathcal{D}$, $D_1\subset D$ and $D_1\not = \emptyset$  implies $D_1\in \mathcal{D}.$ In order to avoid trivialities we assume $\cup_{D\in \mathcal{D}}D=V.$  In the literature such a family $\mathcal{D}$ is called  a hypergraph or an abstract simplicial complex  or more simply the generating class of the hierarchical model. We denote by $\Omega_{\mathcal{D}}$ the linear subspace
of  $x\in {\mathbb{R}}^I$  such that there exist functions $ \theta_D\in {\mathbb{R}}^I$ for $D\in \mathcal{D}$ depending only on $i_D$ and such that $x=\sum_{D\in \mathcal{D}}\theta_D$, that is
\begin{eqnarray}
\label{omegad}\nonumber
\Omega_{\cal D}=\{x\in {\mathbb{R}}^I:\;\exists\theta_D\in {\mathbb{R}}^I, D\in {\cal D}\;\mbox{such that}\; \theta_D(i)=\theta_D(i_D)\;\mbox{and}\; x=\sum_{D\in {\cal D}}\theta_D\}
\end{eqnarray} 
The hierarchical  model generated by ${\cal D}$ is the set of probabilities $p=(p(i))_{i\in I}$ on $I$ such that $p(i)>0$ for all $i$ and such that $\log p\in \Omega_{\cal D}.$

The class of discrete graphical models is a subclass of the class of hierarchical discrete loglinear models. Indeed, let $G=(V, E)$ be an undirected graph where $V$ is the set of vertices and $E\subset V\times V$ denotes the set of undirected edges. We say that the distribution of $X$ is Markov with respect to $G$ if $(v_1,v_2)\not \in E$ implies
$$X_{v_1}\perp X_{v_2}|\; X_{V\setminus \{v_1,v_2\}}.$$
A subset $D$ of $V$ is said to be a clique if for any $v_1,v_2$ in $D$, the edge $(v_1,v_2)$ is in $E$. A clique is said to be maximal if it is maximal with respect to inclusion.
 Let ${\cal D}$ be the set of all cliques (not necessarily maximal) of the graph $G$.
If the distribution of $X=(X_1,\ldots,X_p)$ is Multinomial($1, p(i), i\in I$)  Markov with respect to the graph $G$. By the Hammersley-Clifford theorem, $\log p( i)$ is a linear function of parameters dependent on the marginal cells $i_D, D\in {\cal D}$ only and therefore the graphical model is a hierarchical model with generating set the set ${\cal D}$ of cliques of $G$. The reader is referred to Darroch and Speed (1983), Lauritzen (1996) or Letac \& Massam (2012) for a detailed description of the loglinear model.

In the case of binary data, i.e., when $I_v=\{0,1\}$, for a given ${\cal D}$, the density of the multinomial hierarchical model is traditionally written in the machine learning literature as
\be \label{cumbersome}q(x_1,\ldots,x_p)=\exp \{\sum_{D\in {\cal D}}\langle \theta_D, {\bf 1}_D(x_D)\rangle- k(\theta)\}\ee
where  for $L\in \prod_{v\in D}(I_v\setminus \{0\}),$
$${\bf 1}_{D;L}(x_D):=\prod_{v\in D}x_v=\Big\{\begin{array}{cc}1&\mbox{if}\;x_D=L\\0&\mbox{otherwise},\end{array}$$
where $\langle \theta_D, {\bf 1}_D(x_D)\rangle=\sum_{L\in \{0,1\}^{|D|}}\theta_{D;L} {\bf 1}_{D;L}(x_D)$
 and 
$$k(\theta)=\log \Big (\sum_{x\in \{0,1\}^p}\exp \sum_{D\in {\cal D}}\sum_{x_L\in \{0,1\}^{|D|}}\theta_{D;L}, {\bf 1}_{D;L}(x_D)\Big).$$
This notation is too cumbersome for our purposes and we adopt rather the notation of Letac and Massam (2012) which is extremely simple whether $X_v$ takes two or more values. We now recall this notation and some basic results  for discrete graphical loglinear models.

Among all the values that $X_v$ can take in $I_v, v\in V$, we call one of them $0$. For a cell $i\in I$, we define its support $S(i)$  as 
 $$S(i)=\{v\in V\ ; \ i_{v}\neq 0\}$$
and we define also the following subset  $J$ of $I$  
 \be \label{j} J=\{ j\in I,\ \ S(j)\in \mathcal{D}\}.\ee 
 In the sequel we will call this set the $J$-set of the model.
 For $i\in I$ and $j\in J$, we define the symbol
 $$j\tl i$$
  to mean that $S(j)$ is contained in $S(i)$ and that $j_{S(j)}=i_{S(j)}.$  The relation   $\tl$ has the property that if $j,j'\in J$ and $i\in I$ then 
 \begin{equation}\label{TL}j\tl j'\ \ \mbox{and}\ \ j'\tl i\Rightarrow j\tl i.\end{equation}
 With this notation, in Proposition 2.1 of Letac and Massam (2012), it is shown  that for $i\not \in J,\;\theta_i=0$ and that
 \begin{eqnarray}
 \label{thetaj}
 \theta_j&=&\sum_{j'\in J,\; j'\tl j}(-1)^{|S(j)|-|S(j')|}\log \frac{p(j')}{p(0)},\;j\in J\\
\label{pi}  \log p(i)&=&\theta_0+\sum_{j\in J, j\tl i} \theta_j,\;i\in I\\
 \label{p0} \log p(0)&=&\theta_0.
 \end{eqnarray}
 One then readily derives the density of the multinomial M$(N, p(i), i\in I)$ of the cell counts $\underline{n}=(n(i), i\in I)$, Markov with respect to $G$ to be
 \be \label{base} f(n(i),i\in I)=\exp \langle \theta, \underline{n}\rangle -N k(\theta) \ee
 where 
 $$\langle \theta, \underline{n}\rangle=\sum_{j\in J}\theta_jn(j_{S(j)}), \;\;k(\theta)=\log p(0)^{-1}=\log \Big(\sum_{i\in I}\exp \sum_{j\in J, j\tl i} \theta_j\Big).$$
 Clearly exponential family \eqref{base} for $\theta\in R^{|J|}$ is of dimension $|J|$ and, up to a multiplicative constant,  it is the set of distributions of $t=(n(j_{S(j)}), j\in J)$ with density 
 \begin{equation}
 \label{ftheta}
 f(t;\;\theta)= \exp \{\langle t, \theta \rangle-Nk(\theta)\},\; \theta\in R^{|J|}
 \end{equation}
 with $\theta=(\theta_j,j\in J),\; t=(t(j_{S(j)},\; j\in J$ and $k(\theta)=N\log \Big(\sum_{i\in I} \exp \sum_{j\in J, j\tl i} \theta_j\Big)$.
 
 We note that in the case where $I_v=\{0,1\}$, \eqref{base} is, of course, identical to \eqref{cumbersome}.

\section{The local marginal likelihood estimator}
We assume that $X=(X_v,v\in V)$ follows a discrete graphical model with underlying graph $G$. For a given vertex $v\in V$, let $\N_v$ the set of neighbours of $v$ in the given graph $G$. We write $$\M_v=\{v\}\cup \N_v.$$We first consider  the $\M_v$-marginal model  derived from \eqref{base}. We will see that the corresponding optimization problem is not convex in $\theta$. Following Meng, Wei, Wiesel and Hero (2013), we  define a one-hop and a two-hop convex relaxation of the marginal model and then derive the asymptotic properties of those estimates.
\subsection{The $\M_v$-marginal models}
The $\M_v$-marginal model for $X_{\M_v}$  is clearly multinomial  and the corresponding data  can be read in the $\M_v$-marginal contingency table obtained from the full table. Since in general, the model is not collapsible onto the graph induced from $G$ by $\M_v$, we will first have to derive the generating set  ${\cal D}^{\M_v}$ of the marginal model.
Let $J^{\M_v}$ be the $J$-set of non zero loglinear parameters of the $\M_v$-marginal model, that is
$$J^{\M_v}=\{i\in \I_{\M_v}:\; S(i)\in {\cal D}^{\M_v}\}.$$
The corresponding canonical  parameter is therefore
$$\theta^{\M_v}=(\theta_j^{\M_v}, \;j\in J^{\M_v}).$$
The marginal distribution of $X_{\M_v}$ has density of the form
\begin{equation}
\label{marginal}
f(t^{\M_v};\;\theta^{\M_v})=\exp\{ \langle t^{\M_v}, \theta^{\M_v}\rangle -Nk^{\M_v}(\theta^{\M_v})\}
\end{equation}
where $ t^{\M_v}$ and $k^{\M_v}$ are respectively the canonical statistic and the cumulant generating function of the marginal multinomial density obtained from the density \eqref{ftheta} of $X$.

In order to identify the $\M_v$-marginal model, we first establish the relationship between $\theta$ and $\theta^{\M_v}$. In the sequel, the symbol $j$ will be understood to be an element of $J^{\M_v}$ when used in the notation $\theta_j^{\M_v}$ while it will be understood to be the element of $J$ obtained by padding it with entries $j_{V\setminus \M_v}=0$ when used in the  expression $\theta_j$. It will also be understood that $\theta_j=0$ if $j\not\in J$ and $\theta_j^{\M_v}=0$ when $j\not \in J^{\M_v}$.
We now give the general relationship between the parameters of the overall model and those of the $\M_v$-marginal model.
\begin{lemma}
\label{one}
For $j\in I$, the parameter $\theta_j$ of the overall model and the parameter $\theta_j^{\M_v}$ of the marginal model are linked by the following:
\begin{eqnarray}
\label{main}
 \theta_{j}^{\M_v}
 &=&\theta_{j}+\sum_{j'\;|\;j'\tl_0 j}(-1)^{|S(j)-S(j')|}\log \Big(1+\sum_{i\in \I,\; i_{\N_v}=j'}\exp \sum_{{k\;|\;k\tl i \atop k\not \tl j'} }\theta_{k} \Big)\;.
\end{eqnarray}
\end{lemma}
\noindent {\bf Proof.}
We will  use the notation $j\tl_0 j'$ to mean that $j\tl j'$ or  $j=0$, the zero cell.
By definition \eqref{thetaj} of the loglinear parameters and denoting $p^{\M_v}(j)$ the marginal probability of $j\in I_{\M_v}$, we have 
  \be\label{thetajnv}
 \theta_j^{\M_v}=\sum_{j'\in J^{\N_v},\; j'\tl j}(-1)^{|S(j)|-|S(j')|}\log \frac{p^{\M_v}(j')}{p^{\M_v}(0)}.
 \ee
Also,
\begin{eqnarray*}
p^{\M_v}(j)&=&\sum_{i\in \I.\; i_{\N_v}=j}p(i)
=   \sum_{i\in \I,\; i_{\N_v}=j}\exp \{\sum_{j'\;|\;j'\tl_0 j}\theta_{j'}+\sum_{{j'\;|\;j'\tl i \atop j'\not \tl j} \atop j'_{\N_v}\tl_0 j}\theta_{j'} \}              \\
&=&\Big(\exp \sum_{j'\;|\;j'\tl_0 j}\theta_{j'}\Big)\Big(1+\sum_{i\in \I,\; i_{\N_v}=j}\exp \sum_{{j'\;|\;j'\tl i \atop j'\not \tl j} \atop j'_{\N_v}\tl_0 j}\theta_{j'} \Big)\;.
\end{eqnarray*}
Therefore 
\begin{eqnarray*}
\log p^{\M_v}(j)&=& \sum_{j'\;|\;j'\tl_0 j}\theta_{j'}+\log \Big(1+\sum_{i\in \I,\; i_{\N_v}=j}\exp \sum_{{j'\;|\;j'\tl i \atop j'\not \tl j} }\theta_{j'} \Big)\;,
\end{eqnarray*}
which we can write
\begin{eqnarray}
\label{oneway}
 \sum_{j'\;|\;j'\tl_0 j}\theta_{j'}&=&\log p^{\M_v}(j)-\log \Big(1+\sum_{i\in \I,\; i_{\N_v}=j}\exp \sum_{{k\;|\;k\tl i \atop k\not \tl j} }\theta_{k} \Big)\;.
\end{eqnarray}
Moebius inversion formula states that for $a\subseteq V$ and equality of the form
$\sum_{b\subseteq a}\Phi(b)=\Psi(a)$ is equivalent to $\Phi(a)=\sum_{b\subseteq a}(-1)^{|a\setminus b|}\Psi(b)$. Here, using a generalization of the Moebius inversion formula to the partially ordered set given by $\tl$ on $I$, we derive from  \eqref{oneway} that for $j\in J^{\N_v}\subset J$
\begin{eqnarray}
 \theta_{j}&=&\sum_{j'\;|\;j'\tl_0 j}(-1)^{|S(j)-S(j')|}\log p^{\N_v}(j')-\sum_{j'\;|\;j'\tl_0 j}(-1)^{|S(j)-S(j')|}\log \Big(1+\sum_{i\in \I,\; i_{\N_v}=j'}\exp \sum_{{k\;|\;k\tl i \atop k\not \tl j'} }\theta_{k} \Big)\nonumber\\
 &=&\theta_{j}^{\N_v}-\sum_{j'\;|\;j'\tl_0 j}(-1)^{|S(j)-S(j')|}\log \Big(1+\sum_{i\in \I,\; i_{\N_v}=j'}\exp \sum_{{k\;|\;k\tl i \atop k\not \tl j'} }\theta_{k} \Big)\label{jprime}
\end{eqnarray}
which we prefer to write as \eqref{main}.
$\square$

We now want to identify which of the marginal parameters are equal to the corresponding overall parameter and in particular which ones are equal to $0$. Let $\M_v^c$ denote the complement of $\M_v$ in $V$. We define the buffer set  at $v$ as follows:
$$\B_v=\{w\in M_v\;|\;\exists w'\in \M^c_v\;\mbox{with}\;(w,w')\in E\}.$$
We have the following result.
\begin{lemma}\label{two}
Let $j\in I$ such that $S(j)\subset \M_v$. Then
\begin{enumerate}
\item[(1.)] if $S(j)\not \subset\B_v$, then $\theta^{\M_v}_{j}=\theta_j$,
\item[(2.)] if $S(j)\subset\B_v$, then in general $\theta^{\M_v}_j\not =\theta_j$, and \eqref{main} holds.
\end{enumerate}
In particular this implies that  
$$\theta_j=0\;\mbox{and}\;S(j)\not \subset\B_v \Rightarrow \theta^{\M_v}_{j}=0\;\mbox{i.e.}\;j\not \in J^{\M_v}$$ 
while if $\theta_j=0$ but $S(j) \subset\B_v$, then $j$ might belong to $J^{\M_v}$.
\end{lemma}
 {\bf Proof: }
Since \eqref{main} is already proved, we need only prove that (1.) holds, i.e., that when $S(j)\not \subseteq\B_v$, the alternating sum on the right-hand side of \eqref{main}  is equal to 0. Since $j\in J$, $S(j)$ is necessarily complete and $j'\tl j$ is obtained by removing one or more vertices from $S(j)$.

If $S(j)\cap \B_v=\emptyset$ since any $i\in \I$ such that $i_{\M_v}=j_{\M_v}$ is such that $i_w=0, w\in \M_v,w\not\in S(j)$, the $\theta_k, k\tl i, k\not \tl j'$ in \eqref{main} are the same for all $j'\tl j$. So, the log terms in the alternating sum of logarithms in \eqref{main} are identical and cancel each other out and therefore $\theta^{\M_v}_{j_{\M_v}}=\theta_j$.

If $S(j)\cap \B_v\not =\emptyset$ and $S(j)\not \subseteq \B_v$, there is at least one vertex $w_1\in S(j)$ which is not linked to any point  of the buffer $\B_v$. Let $l_0$  and $l_{w_1}$ be the log terms in the alternating sum corresponding to $j'=0$ and $j'$ such that $S(j')=\{w_1\}$ respectively.
 Since the values of $j_l$ for $l$ a neighbour of $w$ not in $S(j')$ is $0$, for any $i$ such that $i_{\M_v}=j_{\M_v}$, the $\theta_k, k\tl i, k\not \tl j'$ in \eqref{main} are the same for $j'=0$ and $j'$ such that $S(j')=\{w_1\}$. 
 The terms in $l_0$ and $l_w$ are therefore exactly the same except for their sign and these two terms  cancel out. 
 Similarly, for any  $j'\tl j$ such that $\emptyset\not =S(j')\subseteq S(j)\cap \B_v$, $l_{S(j')}$ cancels out with $l_{S(j')\cup \{w_1\}}$. If there is only  $w_1\in S(j)\cap \B_v$, this proves (1.). Otherwise, there is a $w_2\in S(j)\cap \B_v^c, w_2\not =w_1$ and to any subset $w_1\cup S(j'), \;j'\tl j$, we associate the subset $\{w_1, w_2\}\cup S(j')$ and it is clear that the logarithms term $l_{w_1\cup S(j')}$ and $l_{\{w_1, w_2\}\cup S(j')}$ cancel each other out. We iterate this argument until we have taken into account all $w_i\in S(j)\cap \B_v^c$ and (1.) is proven.
$\square$
\vspace{2mm}

\noindent Before proceeding any further, we give an example of the relationship between $\theta$ and $\theta^{\M_v}$.

 \noindent {\bf Example 2} Let $G$ be the 4-cycle with edges $\{(1,2), (1,3), (2,4), (3,4)\}$. Let $v=1$ so that $\N_v=\{1,2,3\}$. Assume the data is binary, taking values in $\{0,1\}$. Let $j^{N_v}=(101)\in J^{\N_v}$ and let us apply the formula above to find the relationship between
$\theta_{101}^{\N_v}$ and $\theta$. Since the $j'\tl_0 j$ are $(000), (100),(001),(101)$, we have
\begin{eqnarray*}
 \theta_{101}^{\N_v}
 &=&\theta_{1010}+\mbox{term for }j'=(000)-\mbox{term for }j'=(100)-\mbox{term for }j'=(001)\\
 &&\hspace{4cm}+\mbox{term for }j'=(101)
\end{eqnarray*}
where $j'$ refers to the running index in the right hand side of \eqref{jprime}.

\noindent For $j'=(000),$ the $i\in \I$ such that $i_{\N_v}=j'$ are $(0000)$ and $(0001)$. In each case, the set of $k\in J$ such that $k\tl i, k\not \tl j', k_{\N_v}\tl_0 j'$ is $\emptyset$ and $\{(0001)\}$, respectively.
\newline For $j'=(100),$ the $i\in \I$ such that $i_{\N_v}=j'$ are $(1000)$ and $(1001)$. In each case, the set of $k\in J$ such that $k\tl i, k\not \tl j', k_{\N_v}\tl_0 j'$ is $\emptyset$ and $\{(0001)\}$, respectively.
\newline For $j'=(001),$ the $i\in \I$ such that $i_{\N_v}=j'$ are $(0010)$ and $(0011)$. In each case, the set of $k\in J$ such that $k\tl i, k\not \tl j', k_{\N_v}\tl_0 j'$ is $\emptyset$ and $\{(0011)\}$, respectively.
\newline For $j'=(101),$ the $i\in \I$ such that $i_{\N_v}=j'$ are $(1010)$ and $(1011)$. In each case, the set of $k\in J$ such that $k\tl i, k\not \tl j', k_{\N_v}\tl_0 j'$ is $\emptyset$ and $\{(1011)\}$, respectively.
Formula \eqref{jprime} therefore yields
\begin{eqnarray*}
 \theta_{(101)}^{\N_v} &=&\theta_{(1010)}+\log (1+\exp \theta_{(0001)})-\log (1+\exp \theta_{(0001)})-\log (1+\exp (\theta_{(0001)}+\theta_{(0011)}))\\
 &&+\log (1+\exp (\theta_{(0001)}+\theta_{(0011)}))\\
 &=&\theta_{(1010)}
\end{eqnarray*}
Following the same procedure, we find
\begin{eqnarray*}
 \theta_{(010)}^{\N_v} &=&\theta_{(0100)}-\log (1+\exp \theta_{(0001)})+\log (1+\exp \theta_{(0001)}+\theta_{(0101)}).
\end{eqnarray*}
We note that, illustrating Lemma \ref{two},  $\theta_{(101)}^{\N_v} =\theta_{(1010)}$ but $\theta_{(010)}^{\N_v} \not =\theta_{(0100)}$.

\subsection{A convex relaxation of the marginal optimization problems}
It is clear from \eqref{main} that even though maximizing the marginal likelihood from \eqref{marginal}   is  convex in $\theta^{\M_v}$, it is not convex in $\theta$. However, we know from Lemma \ref{two} that
\begin{equation}
\label{mv=o}
\{j\in J^{\M_v}:\;S(j)\not \subset \B_v\}=\{j\in J:\; S(j)\subset \M_v, \;S(j)\not \subset \B_v\}.
\end{equation}
Therefore maximizing \eqref{marginal} in $\theta^{\M_v}$ will yield  estimates of $\theta_j, j\in J, S(j)\not \subset \B_v$    while the estimates of $\theta^{\M_v}_j$ for $S(j)\subset \B_v$ obtained from the marginal likelihood are discarded. From formula \eqref{main}, it is clear that, for $S(j)\subset \B_v$, in general, the parameters $\theta^{\M_v}_j$ are different from $0$. It is then reasonable to consider a ''relaxation'' of the $\M_v$-marginal model by 
\begin{itemize}
\item not assuming any equality-to-$0$ constraints for the parameters $\theta_j, S(j)\subset \B_v$
\item but keeping the constraints \eqref{mv=o} 
\end{itemize}
We call this relaxed marginal problem the $\M_{1,v}$-marginal problem. It is defined by the following $J$-set
\begin{equation}
\label{jm1v}
J^{\M_{1,v}}=\{j\in J|\; S(j)\subset \M_v,\; S(j)\not \subset \B_v\}\cup \{i\in I|\;  S(i)\subset \B_v\} \;.        
\end{equation}
The corresponding canonical parameter and  canonical statistics are denoted $\theta^{\M_{1,v}}$ and $t^{\M_{1,v}}.$

The local estimates of $\theta_j, j\in \{j\in J|\; S(j)\subset \M_v,\; S(j)\not \subset \B_v\}$ are obtained by maximizing the $\M_{1,v}$-marginal loglikelihood 
\begin{eqnarray}
\label{relax}
\mbox{max}_{\theta^{\M_{1,v}}}\exp \langle \theta^{\M_{1,v}},t^{\M_{1,v}}\rangle -Nk^{\M_{1,v}}(\theta^{\M_{1,v}})
\end{eqnarray} 
 At this  point, we need to make an important remark. The $\M_{1,v}$-marginal model is a hierarchical model but not necessarily a graphical model. For example, if we consider a four-neighbour lattice and a given vertex $v_0$ and its four neighbours that we will call $1,2,3,4$ for now, then the generating set of the relaxed $\M_{1,v_0}$-marginal model is
 $${\cal D}^{\M_{1,v_0}}=\{(v_0,1),  (v_0,2), (v_0,3), (v_0,4), (1,2,3,4)\}.$$
 This is not a discrete graphical model since a graphical model would also include the interactions $(v_0,1,2),(v_0,2,3),(v_0,3,4),(v_0,1,4),(v_0,1,2,3,4).$
 It was therefore crucial to set up our problem, as we did it in Section 2, within the framework of hierarchical loglinear models rather than the more restrictive class of graphical models.
 \vspace{2mm}

\noindent {\bf The local marginal likelihood estimates.}
 From each  maximization problem \eqref{relax} at $v\in V$, we will obtain an estimate   $\hat{\theta}_{j}^{\M_{1,v}}$ for $ j\in \{j\in J|\; S(j)\subset \M_v,\; S(j)\not \subset \B_v\}$.  
 Since for any $j\in J$, $S(j)$ is complete, an estimate of $\theta_j$ will be obtained from each $v\in S(j)$ such that $S(j)\not \subset \B_v$.
 The global estimate of $\theta_j$ can then be derived in various ways as described in  Liu and Ihler (2013). This includes linear consensus, maximum consensus or ADMM. These require extra calculations and one might wonder if we could, as done in Meng, Wei, Wiesel and Hero (2013) define the two-hop marginal estimate and see if the level of accuracy of the local estimates is then such that one does not need to use either consensus or ADMM to obtain an accurate estimator of $\theta_j, j\in J$. We therefore now look at the two-hop marginal estimate and the asymptotic properties of both the local $\M_{1,v}$ and $\M_{2,v}$-marginal estimates.
 \vspace{2mm}

\noindent {\bf The two-hop relaxed marginal model}
For a given $v\in V$, we  now modify  the definition of $\N_v$ to include not only the neighbours of $v$ in $G$ but also the neighbours of the neighbours. Let $\M_v=\{v\}\cup \N_v$ and $\B_v=\{w\in \M_v|\;\exists w'\in \M_v^c, \;(w,w')\in E\}$. Lemmas \ref{one} and \ref{two} also hold for this enlarged neighbourhood and we define the two-hop relaxed $\M_{2,v}$-marginal neighbourhood to be the model with $J$-set equal to 
 \begin{equation}
\label{jm2v}
J^{\M_{2,v}}=\{j\in J|\; S(j)\subset \M_v,\; S(j)\not \subset \B_v\}\cup \{i\in I|\;  S(i)\subset \B_v\} \;.        
\end{equation}
where $\M_v$ and $\B_v$ are the new neighbourhood and buffer sets.
\section{Properties of the local  $\M_{i,v}$-marginal estimates}
In this section, we first show that for $i=1,2$ the  $\M_{i,v}$-marginal estimates $\hat{\theta}^{\M_{i,v}}_j,\; S(j)\not \subset \B_v$ are asymptotically consistent and then we will show that the asymptotic variance of $\hat{\theta}^{\M_{2,v}}_j$ is always less than or equal to that of $\hat{\theta}^{\M_{1,v}}_j$.
For $i=1,2$, let 
\begin{equation}
\label{miv}
\hat{\theta}^{\M_{i,v}}=\mbox{argmax}_{\theta^{\M_{i,v}}}\; \langle \theta^{\M_{i,v}}, t^{\M_{i,v}}\rangle -Nk^{\M_{i,v}}(\theta^{\M_{i,v}})\;.
\end{equation}
be the mle of the canonical parameter in the relaxed $\M_{i,v}$-model. The following results are valid for $i=1$ and $i=2$.
\begin{lemma}
\label{cont}
For $v\in V$ given, the estimate $\hat{\theta}^{\M_{i,v}}$, as defined in \eqref{miv} above, is  a continuous function of $t^{\M_v}=(n(i_E), \;E\subset \M_v, i\in E)$, the set of all possible marginal counts of the $\M_v$-marginal contingency table.
\end{lemma}
{\bf Proof.} We first recall that the canonical statistic corresponding to any multinomial model  is the  vector of marginal counts corresponding to the $J$-set of that model.
So $t^{\M_{i,v}}$ is a linear function of $t^{\M_v}$. Since we are dealing with natural exponential families, $\hat{\theta}^{\M_{i,v}}$  is given by
$$\hat{\theta}^{\M_{i,v}}=\psi^{\M_{i,v}}(\frac{t^{\M_{i,v}}}{N})$$
where $\psi^{\M_{i,v}}$ is the inverse function of $\theta\mapsto \frac{d k^{\M_{i,v}}(\theta)}{d \theta}$ and is a continuous function.
\newline $\square$
\begin{theorem}
\label{consistency}
The mle $\hat{\theta}$   obtained by concatenating the estimates  $\hat{\theta}^{\M_{i,v}}_j, S(j)\subseteq \M_v, S(j)\not \subseteq \B_v,\;v\in V$, obtained  from \eqref{miv} for $ v\in V$, is asymptotically consistent.
\end{theorem}
{\bf Proof: } Let $\theta_*$ be the true parameter for model \eqref{ftheta} and let  $\theta_{* }^{\M_v}$ be the corresponding true parameter for the marginal model \eqref{marginal}.  Let $p_*^{\M_v}$ be the corresponding vector of true marginal cell probabilities for \eqref{marginal} and let $t_*^{\M_v}=Np_*^{\M_v}$. The first step is to show that $\theta_{* }^{\M_v}$ is the optimal solution of the relaxed marginal optimization problem \eqref{relax} when the canonical statistic in \eqref{relax} is equal to  $t_*^{\M_v}$.

To do so, we first note that $\theta_{* }^{\M_v}$ is feasible for \eqref{relax}, i.e., it belongs to $J^{\M_{i,v}}$. This follows from Lemma \ref{cont} and the definition of $J^{\M_{i,v}}$. Next, we verify that it satisfies the optimality condition of the $\M_v$-marginal model which which is
$$\theta_{* }^{\M_v}=\psi^{\M_{i,v}}(t_*^{\M_v})$$
or equivalently
\begin{equation}
\label{opt}
t_*^{\M_v}=Np_*^{\M_v}=N\frac{d\; k^{\M_{i,v}}(\theta^{\M_{i,v}})}{d\; \theta^{\M_{i,v}}}{\Big|_{\theta^{\M_{i,v}}=\theta_{*}^{\M_v}}}
\end{equation}
To see that the equality above is true, first we recall that in natural exponential families with cumulant generating function $k(\theta)$, the mean parameter is given by $\frac {d\;k(\theta)}{d\; \theta}$. Next we compare the cumulant generating functions of the two models \eqref{marginal} and \eqref{miv} which are respectively
\begin{eqnarray*}
k^{\M_v}=\log\Big( \sum_{i\in I_{\M_v}}\exp \sum_{j\in J^{\M_v}|\;j\tl i}\theta_j\Big)\label{nonrel}\\
k^{\M_{i,v}}=\log\Big( \sum_{i\in I_{\M_{i,v}}}\exp \sum_{j\in J^{\M_{i,v}}|\;j\tl i}\theta_j\Big)\label{rel}
\end{eqnarray*}
Since $J^{\M_v}\subset J^{\M_{i,v}}$, we see  immediately that the derivative of $k^{\M_{i,v}}$ evaluated at
$\theta^{\M_{i,v}}=\theta_{*}^{\M_v}$ is equal to the derivative of $k^{\M_v}$ at $\theta_{*}^{\M_v}$. Therefore \eqref{opt} is satisfied and we can conclude that $\theta_{* }^{\M_v}$ is the unique optimal solution of \eqref{relax} when the canonical statistic is equal to  $t_*^{\M_v}$.

For the second step in our proof, we recall that as $N\rightarrow +\infty$, by the law of large numbers, $\frac{t^{\M_v}}{N}\rightarrow p^{\M_v}_*$. Therefore, as $N\rightarrow +\infty$, by  Lemma \ref{cont} and the continuous mapping theorem, the maximum likelihood estimate $\hat{\theta}^{\M_{i,v}}$ of \eqref{relax} for $t^{\M_{i,v}}$ arbitrary tends to the maximum likelihood estimate  of \eqref{relax} for $t^{\M_{i,v}}=t^{\M_v}_*$ which is $\theta^{\M_v}_*$.

Finally, since, by Lemma \ref{two},  for $j\in J, S(j)\subset \M_v, S(j)\not \subset \B_v,\; (\theta^{\M_v}_*)_j=(\theta_*)_j$, the asymptotic consistency of the global estimate obtained by concatenating the estimates $\theta^{\M_{i,v}}_j,\;j\in J, S(j)\subset \M_v, S(j)\not \subset \B_v$ is proved. 
$\square$
\vspace{2mm}

We now examine the asymptotic variance of the two estimates $\theta^{\M_{i,v}}_j,\;i=1,2,\;j\in J, S(j)\subset \M_v, S(j)\not \subset \B_v$.
We will now distinguish between the buffer set of the relaxed $\M_{1,v}$-marginal model and that of the $\M_{2,v}$-marginal model  and denote them $\B_{i,v}, i=1,2$ respectively.
We will use the notation
\begin{eqnarray*}
J_{i,v}&=&\{i\in \I_{\M_{i,v}}: \;i\in J, \;S(i)\subset \M_{i,v}, \;S(i)\not \subset \B_{i,v}\}\subset J^{\M_{i,v}}\\
B_{i,v}&=&\{i\in \I_{\M_{i,v}}:\; S(i)\subset \B_{i,v}\}\\
\theta_{J_{i,v}}&=&(\theta_j:\; j\in J_{i,v})\\
\theta_{B_{i,v}}&=&\{\theta_j, j\in B_{i,v}\}
\end{eqnarray*}
We will consider the following four models that are defined by their $J$-sets that we temporarily denote ${\cal J}$:
\begin{enumerate}
\item the relaxed one-hop marginal model $\M_{1,v}$ with $J$-set equal to ${\cal J}=J_{1,v}\cup B_{1,v}$,
\item the relaxed one-hop marginal model $\M_{2,v}$ with $J$-set equal to ${\cal J}=J_{2,v}\cup B_{2,v}$,
\item the overall model with $J$-set ${\cal J}=J$
\item a new augmented marginal model, denoted $\bar{\M}_{2,v}$ that we will use in the argument below with $J$-set equal to ${\cal J}=J_{1,v}\cup B_{1,v}\cup J_{2\setminus 1,v}\cup B_{2,v}$ where $J_{2\setminus 1,v}=J_{2,v}\setminus J_{1,v}$.
\end{enumerate}
We note that the density of the four models is of the general form \eqref{ftheta} with $\theta=(\theta_j, j\in {\cal J})$ and with cumulant generating functions
\begin{eqnarray*}
k^{\M_{i,v}}(\theta^{\M_{i,v}})&=&\log(\sum_{k\in \I_{\M_{i,v}}}e^{\sum_{j\triangleleft k,\;j\in {\cal J}}\theta_j})\\
k^{J}(\theta)&=&\log (\sum_{i\in \I}e^{\sum_{j\triangleleft i,\;j\in {\cal J}}\theta_j})\\
k^{\bar{\M}_{2,v}}(\theta^{\bar{\M}_{2,v}})&=&\log (\sum_{k\in \I_{\M_v}}e^{\sum_{j\triangleleft k,\;j\in {\cal J}}\theta_j})
\end{eqnarray*}
for the  models $\M_{i,v},\;i=1,2$, the overall model and the augmented marginal model $\bar{\M}_{2,v}$ respectively and where the set ${\cal J}$ changes accordingly.

Whatever the model, the symmetric matrix of the covariance of $t$ is the ${\cal J}\times {\cal J}$ matrix
\begin{eqnarray*}
\frac{\partial^2k(\theta)}{\partial \theta^2}&=&\Big(\frac{\partial^2k(\theta)}{\partial \theta_j \partial \theta_{j'}}\Big)_{j,j'\in \cal J}=
\Big(p_{j\cup j'}-p_jp_{j'}\Big)_{j,j'\in \cal J}
\end{eqnarray*}
where we use the notation $j\cup j'$ to denote the cell $i\in  \I_{\M_{i,v}}$ or $i\in \I$ with support $j\cup j'$ and
$$p_{j\cup j'}=p((j\cup j')_{S({j\cup j'})}),\;\;p_j=p(j_{S(j)})$$
denote marginal probabilities. For $j, j'$ given, since $p_{j\cup j'}, p_j, p_{j'}$  are marginal probabilities the entries $p_{j\cup j'}-p_jp_{j'}$ are the same  for all models with $j,j'\in {\cal J}$.
We will now prove the following result concerning the variance of the estimates.
\begin{theorem}
\label{variance}
Let $v\in V$ be given. For $j\in J_{1,v}$, the local maximum likelihood estimates $\hat{\theta}_{j}^{\M_{i,v}}, \;i=1,2$ obtained from \eqref{miv} and the maximum likelihood estimate $\hat{\theta}_j$ obtained from the overall  $J$ model are such that
\begin{eqnarray}
\label{var}
\mbox{var}(\hat{\theta}_{j}^{1,v})\geq \mbox{var}(\hat{\theta}_{j}^{2,v})\geq \mbox{var}(\hat{\theta}_j).
\end{eqnarray}
\end{theorem}

 {\bf Proof:}
From standard asymptotic theory, we know that the asymptotic variance of $\theta^{\bar{\M}_{2,v}}$ is equal to 
\begin{equation}
\Big(\frac{\partial^2 k^{\bar{\M}_{2,v}}(\theta^{\bar{\M}_{2,v}})}{\partial (\theta^{\bar{\M}_{2,v}})^2}\Big)^{-1}
\end{equation}
evaluated at the corresponding true value of the parameter. It will be convenient in the sequel to represent the symmetric matrix $K=\frac{\partial^2 k^{\bar{\M}_{2,v}}(\theta^{\bar{\M}_{2,v}})}{\partial (\theta^{\bar{\M}_{2,v}})^2}$ according to the different blocks determined by the subvectors of $\theta^{\bar{\M}_{2,v}})$ as follows
$$K=\left(\begin{array}{cccc}K_{J_{1,v},J_{1,v}}&K_{J_{1,v},B_{1,v}}&K_{J_{1,v},J_{2\setminus 1,v}}&K_{J_{1,v},B_{2,v}}\\
K_{B_{1,v}, J_{1,v}}&K_{B_{1,v}, B_{1,v}}&K_{B_{1,v}, J_{2\setminus 1,v}}&K_{B_{1,v}, B_{2,v}}\\
K_{ J_{2\setminus 1,v},J_{1,v}}&K_{ J_{2\setminus 1,v},B_{1,v}}&K_{ J_{2\setminus 1,v},J_{2\setminus 1,v}}&K_{ J_{2\setminus 1,v},B_{2,v}}\\
K_{B_{2,v}, J_{1,v}}&K_{B_{2,v}, B_{1,v}}&K_{B_{2,v}, J_{2\setminus 1,v}}&K_{B_{2,v}, B_{2,v}}
\end{array}
\right).$$

We observe that in the $\bar{\M}^{2,v}$ model, the subset $\B_{1,v}\subset V$ separates $\{v\}$ from $V\setminus \M_{1,v}$ and the set $\B_{1,v}$ is complete. Therefore using a standard formula in graphical models, we have that
\begin{eqnarray*}
K^{-1}&=&\left(\begin{array}{cc}K_{J_{1,v},J_{1,v}}&K_{J_{1,v},B_{1,v}}\\K_{B_{1,v}, J_{1,v}}&K_{B_{1,v}, B_{1,v}}\end{array}\right)^{-1}
+\left(\begin{array}{ccc}K_{B_{1,v}, B_{1,v}}&K_{B_{1,v}, J_{2\setminus 1,v}}&K_{B_{1,v}, B_{2,v}}\\
K_{ J_{2\setminus 1,v},B_{1,v}}&K_{ J_{2\setminus 1,v},J_{2\setminus 1,v}}&K_{ J_{2\setminus 1,v},B_{2,v}}\\
K_{B_{2,v}, B_{1,v}}&K_{B_{2,v}, J_{2\setminus 1,v}}&K_{B_{2,v}, B_{2,v}}\end{array}\right)^{-1}-K_{B_{1,v}, B_{1,v}}^{-1}
\end{eqnarray*}
where matrices on the right-hand-side of the equation are "padded" with zeros  in the appropriate blocks.

Let $\theta_{J_{1,v}}=(\theta_j, \; j\in J_{1,v})$. Then the covariance matrix of $(\hat{\theta}^{\bar{\M}_{2,v}})_{J_{1,v}}$ is $[K^{-1}]_{J_{1,v}}$. From the previous expression of $K^{-1}$, we have
\begin{eqnarray}
\label{global}
[K^{-1}]_{J_{1,v}}=\left[\left(\begin{array}{cc}K_{J_{1,v},J_{1,v}}&K_{J_{1,v},B_{1,v}}\\K_{B_{1,v}, J_{1,v}}&K_{B_{1,v}, B_{1,v}}\end{array}\right)^{-1}\right]_{J_{1,v}}
\end{eqnarray}
Since $(\theta_j,\;j\in J_{1,v}\cup B_{1,v})=\theta^{\M_{1,v}}$, we have that
$$\left(\begin{array}{cc}K_{J_{1,v},J_{1,v}}&K_{J_{1,v},B_{1,v}}\\K_{B_{1,v}, J_{1,v}}&K_{B_{1,v}, B_{1,v}}\end{array}\right)=\frac{\partial^2 k^{\M_{1,v}}}{\partial( \theta^{\M_{1,v}})^2}=[\mbox{var}(\theta^{\M_{1,v}})]^{-1}
$$ and therefore
\begin{eqnarray}
\label{j1}
[K^{-1}]_{J_{1,v}}=\left[[\mbox{var}(\hat{\theta}^{\M_{1,v}})\right]_{J_{1,v}}=\mbox{var}([\hat{\theta}^{\M_{1,v}}]_{J_{1,v}})\;.
\end{eqnarray}
Moreover, using standard linear algebra formulas, we have that
\begin{eqnarray}
[K^{-1}]_{J_{1,v}}&=&\left(K_{J_{1,v}\bullet (B_{1,v}\cup J_{2\setminus 1,v}\cup B_{2,v} )}\right)^{-1}\geq \left(K_{J_{1,v}\bullet ( J_{2\setminus 1,v}\cup B_{2,v} )}\right)^{-1}
=\left[(K_{J_{1,v}\cup  J_{2\setminus 1,v}\cup B_{2,v} })^{-1}\right]_{J_{1,v}},\nonumber\\
(K_{J_{1,v}\cup  J_{2\setminus 1,v}\cup B_{2,v} })^{-1}&=&\mbox{var}(\hat{\theta}^{\M_{2,v}}),\label{j2}\\
(K_{J_{1,v}\cup  J_{2\setminus 1,v}\cup B_{2,v} })^{-1}&\geq& (K_{J_{1,v}\cup  J_{2\setminus 1,v}\cup (B_{2,v}\cap J) })^{-1}=(K_J)^{-1}=\mbox{var}(\hat{\theta})\label{jj}
\end{eqnarray}
Combining \eqref{global}, \eqref{j1} and \eqref{j2}, we obtain that
$$\mbox{var}([\hat{\theta}^{\M_{1,v}}]_{J_{1,v}})\geq \mbox{var}([\hat{\theta}^{\M_{2,v}}]_{J_{1,v}})$$
which is the first inequality in \eqref{var}. Now, combining \eqref{j2} and \eqref{jj}, we obtain that
$$\mbox{var}([\hat{\theta}^{\M_{2,v}}]_{J_{1,v}})\geq \mbox{var}([\hat{\theta}]_{J_{1,v}})$$
and taking the diagonal elements of those matrices yields \eqref{var}.
 $\square$
 
{\bf Remark:} Meng et al. (2013) state the following: ''In principle, larger neighborhoods would
allow each node to access more data and hence increase its information for estimating its local parameters.''
 This is basically equivalent to saying that the variance of an estimate based on a larger neighbourhood will be smaller  than that of an estimate  based on a smaller neighbourhood. However, the relationship between the variances of the various estimates was not proved for the Gaussian case. Following the proof of Theorem \ref{variance}, it is clear that a parallel proof  will establish the fact that the variance of the two-hop estimate is smaller than that on the one-hop estimate, for the Gaussian case as well.

\section{Numerical experiments}
\subsection{Accuracy of the estimates}
In this section, we compare the numerical values of the estimates of $\theta$ using
\begin{itemize}
\item the local pseudo- likelihood with simple averaging, abbreviated ps-mle,
\item the local one-hop marginal likelihood,
\item the local two-hop marginal likelihood,
\item the global likelihood of the overall model
\end{itemize}
as well as their variances. We consider two $k\times k$  four-neighbour lattice graphs, $k=4$
and $k=10$ .

\begin{figure}[!t]
\begin{center}
\begin{tabular}{cc}
\scalebox{0.4}{\includegraphics{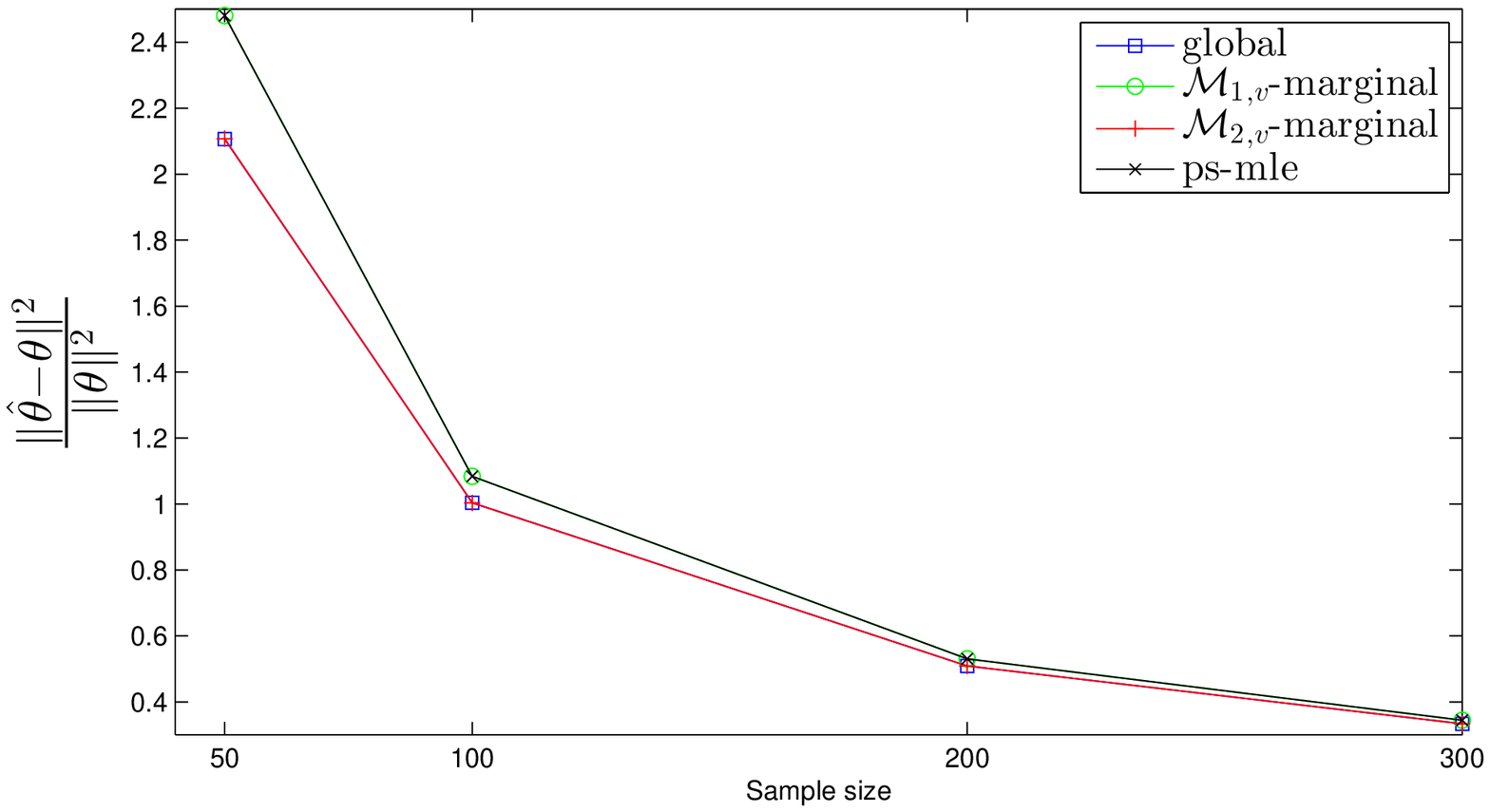}}&
\scalebox{0.4}{\includegraphics{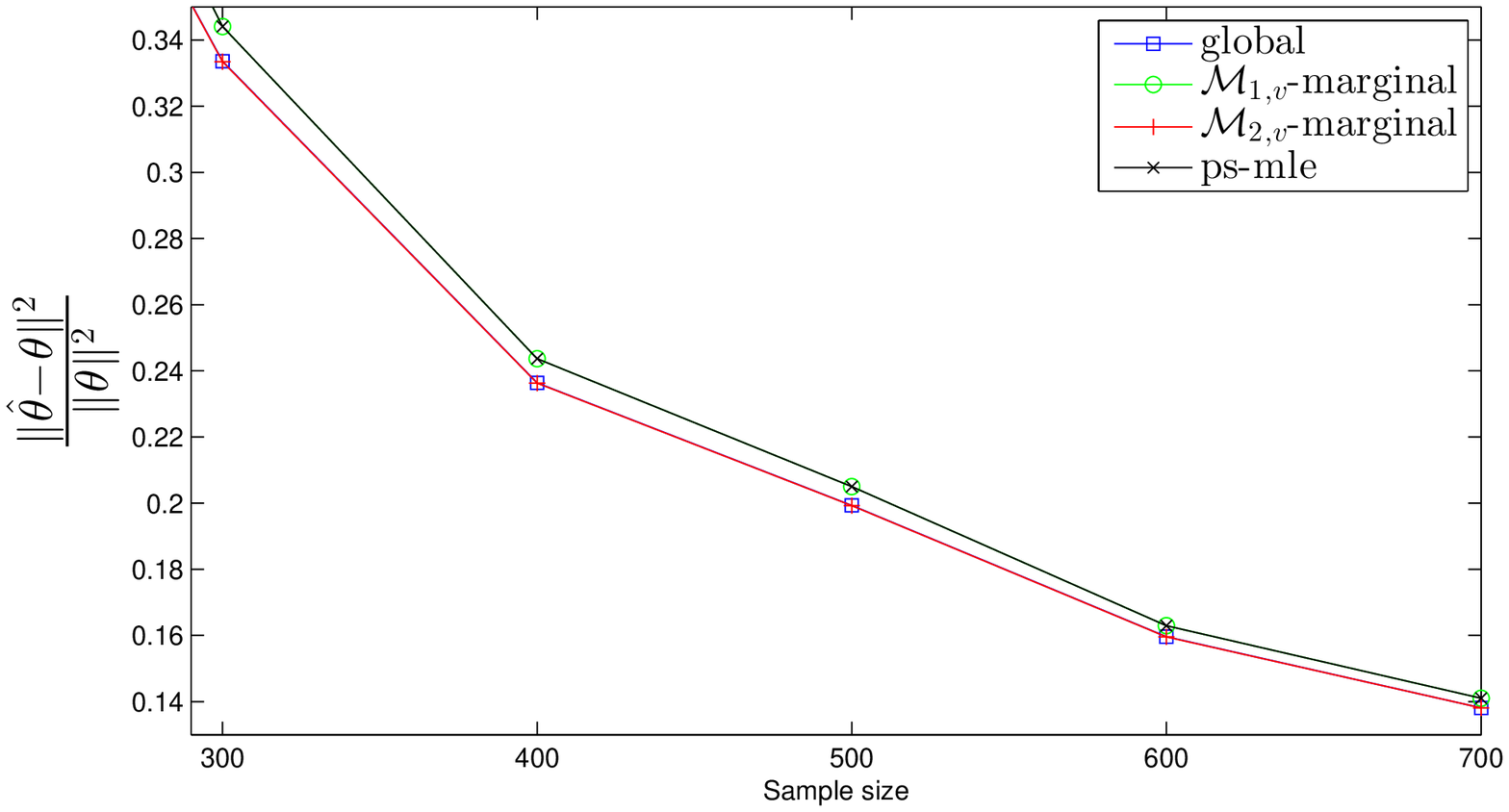}}\\
\end{tabular}
\caption{Plot of the relative mean square error versus sample size, for all four estimates, when the graph is the four-neighbour lattice with $k=4$\label{mse-all}}
\end{center}
\end{figure}

 We generated $n$ samples from each of these two graphical models for $n$ ranging from $50$ to $700$.
For the  small sample size $n=50$, we found that in about 20\% of cases, the local mle's did not exist, whether they were computed using pseudo-likelihood or relaxed marginal likelihoods. Such samples were discarded.  All four estimates were computed (without any consensus or ADMM adjustments) using the remaining samples. The  mean square error was averaged over the remaining experiments. The total number of experiments was 100. These results for mean square   error are illustrated in Figure \ref{mse-all}  for the  four-neighbour lattice with $k=4$, where the relative mean square error
$$\frac{||\hat{\theta}-\theta||^2}{||\theta||^2}= \frac{\sum_{j\in {\cal J}}(\hat{\theta}_j-\theta_j)^2}{\sum_{j\in {\cal J}}\theta_j^2} $$ is plotted versus sample size. The graph is split into two parts so that the difference in error is easier to read for larger sample sizes. 
The pseudo-likelihood estimate was computed using the MATLAB program CVX. The one-hop and two-hop marginal likelihood estimates were computed using the IPF algorithm since it is much faster than maximizing the likelihood function. However, when performing the IPF algorithm, to avoid marginal counts of $0$, we added an extremely small number to each cell count, which for $k= 4, 10$ is respectively of the order of $2^{-30}, 2^{-120}$. These numbers are so small that they do not affect the estimation of the parameters. This was verified numerically in the case $k=4$ by computing the one-hop marginal estimate both using CVX and using IPF.

We  found that the two-hop mle and the global mle were extremely close for $k=4$. We did not compare the two-hop mle and the global mle for $k=10$ since in that case, we could not compute the exact value of the mle. In all cases, the one-hop marginal mle and the pseudo-likelihood estimate are basically numerically equal. 
The times taken to estimates the parameters relative to a given node with four neighbours  in the lattice with $k=4$  and $k=10$ are given in the following table
\vspace{2mm}

\begin{tabular}{|c|c|c|}
\hline
Estimate& time in seconds, $k=4$& time in seconds, $k=10$\\
\hline
$\M_{1,v}$&.5&.5\\
ps-mle&11&11\\
$\M_{2,v}$&50&935\\
\hline
\end{tabular}
\vspace{2mm}

These times are just for finding the estimates without any adjustments such as linear or max consensus or ADMM maximization. We observe that the one-hop marginal method is by far the fastest.The one-hop estimate is faster than the pseudo-likelihood since it uses the IPF algorithm . The two-hop marginal estimate is also obtained using the IPF but the number of cells and cliques increases exponentially with the number of layers in the neighborhood. We still prefer to use the IPF rather than maximizing the relaxed likelihood function since the completion of the buffer introduces a very large number of parameters. So the computation time for the two-hop marginal likelihood estimate will increase with the number of vertices in the two-hop neighborhood but this number  depends only the local topology of the graph. The advantage of the two-hop estimate is clearly its accuracy and therefore the fact that it does not need any further adjustment through consensus or ADMM.

{\bf Remark 1}: The four-neighbour lattice with $k=4$ was considered also in  Liu and Ihler (2012). If we try to compare  the magnitude of the mean square error there and in this paper, for this particular graph, it may look like our errors are larger. This is only due to the fact that the models in the two instances are slightly different in the following sense. In the former, the values taken by the $X_v$ are $\pm1$ while here, $X_v$ takes its values in $\{0,1\}$. This implies that our parameters $\theta_j$ are integer multiples of the corresponding parameters when $X_v=\pm 1$ and the amplitude of the errors is magnified accordingly, even for the relative errors as we computed it. After this is taken into account, the magnitude of the error is roughly of the same order. 
\begin{figure}[htp]
\centering
  \includegraphics[width=15cm]{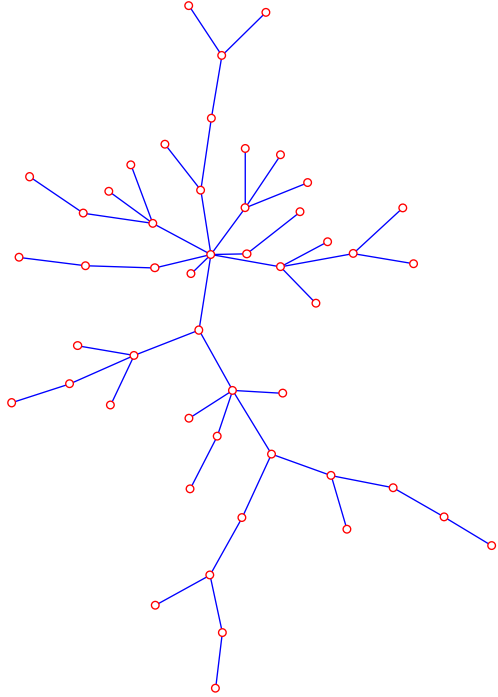}
\caption{ A star graph}
\label{star}
\end{figure}

{\bf Remark 2}: For any star graph $G$ (e.g. Figure \ref{star}), there is no need to do distributed computing since we have an explicit formula for any $\theta_j, j\in J$. Indeed, for any $j\in J$, let ${\cal C}^j, {\cal S}^j$ be the set of cliques $C$ and separators $S$  such that $S(j)\subset C (S(j)\subset S)$ respectively. Let $\theta^C_j,\; \theta^S_j$ be the corresponding parameters in the marginal distribution of $X_C, X_S$ respectively. The following formula 
$$\theta_j=\sum_{C\in {\cal C}^j}\theta^{C}_j-\sum_{S\in {\cal S}^j}\theta^{S}_j$$
can be immediately obtained from the expression of $p(i), i\in I$ in terms of the marginal probabilities of the cliques and separators.
\vspace{4mm}

\subsection{Sample variance of the estimates}
\begin{figure}[htp]
\centering
  \includegraphics[width=15cm]{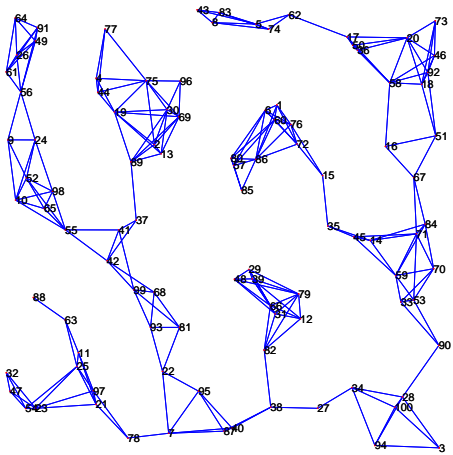}
\caption{ A random graph with 100 vertices}
\label{randomg}
\end{figure}

\begin{figure}[htp]
\centering
  \includegraphics[width=15cm]{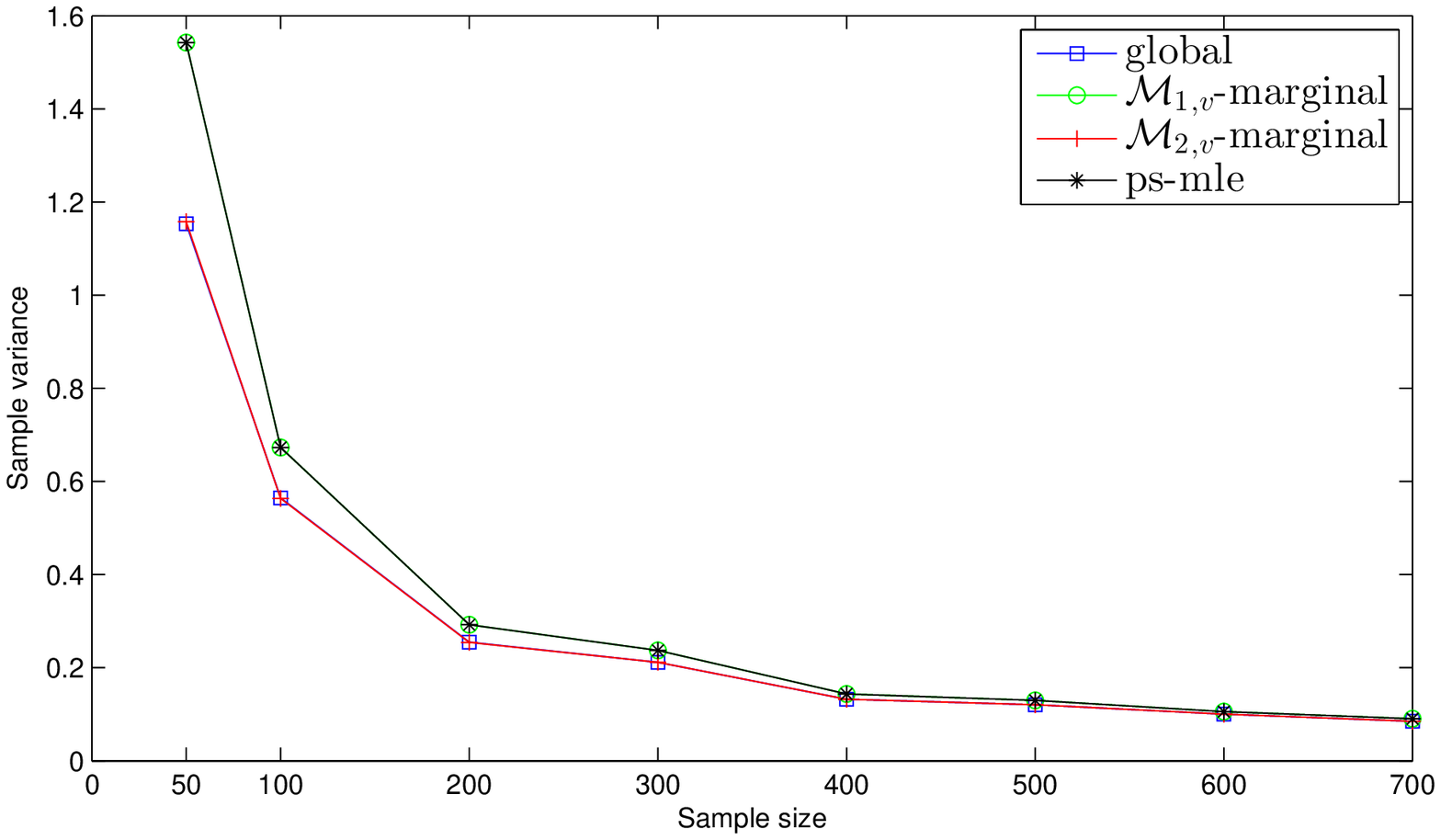}
\caption{Plot of the variance of the relaxed $\M_{i,v},\;i=1,2$- marginal estimates $\hat{\theta}^{\M_{i,v}}_j,\;i=1,2$, the global estimate $\hat{\theta}$ and the pseudo-likelihood estimate $\hat{\theta}^{ps-mle}$ in the four-neighbour $4\times 4$ lattice.}
\label{var7}
\end{figure}

\begin{figure}[htp]
\centering
  \includegraphics[width=15cm]{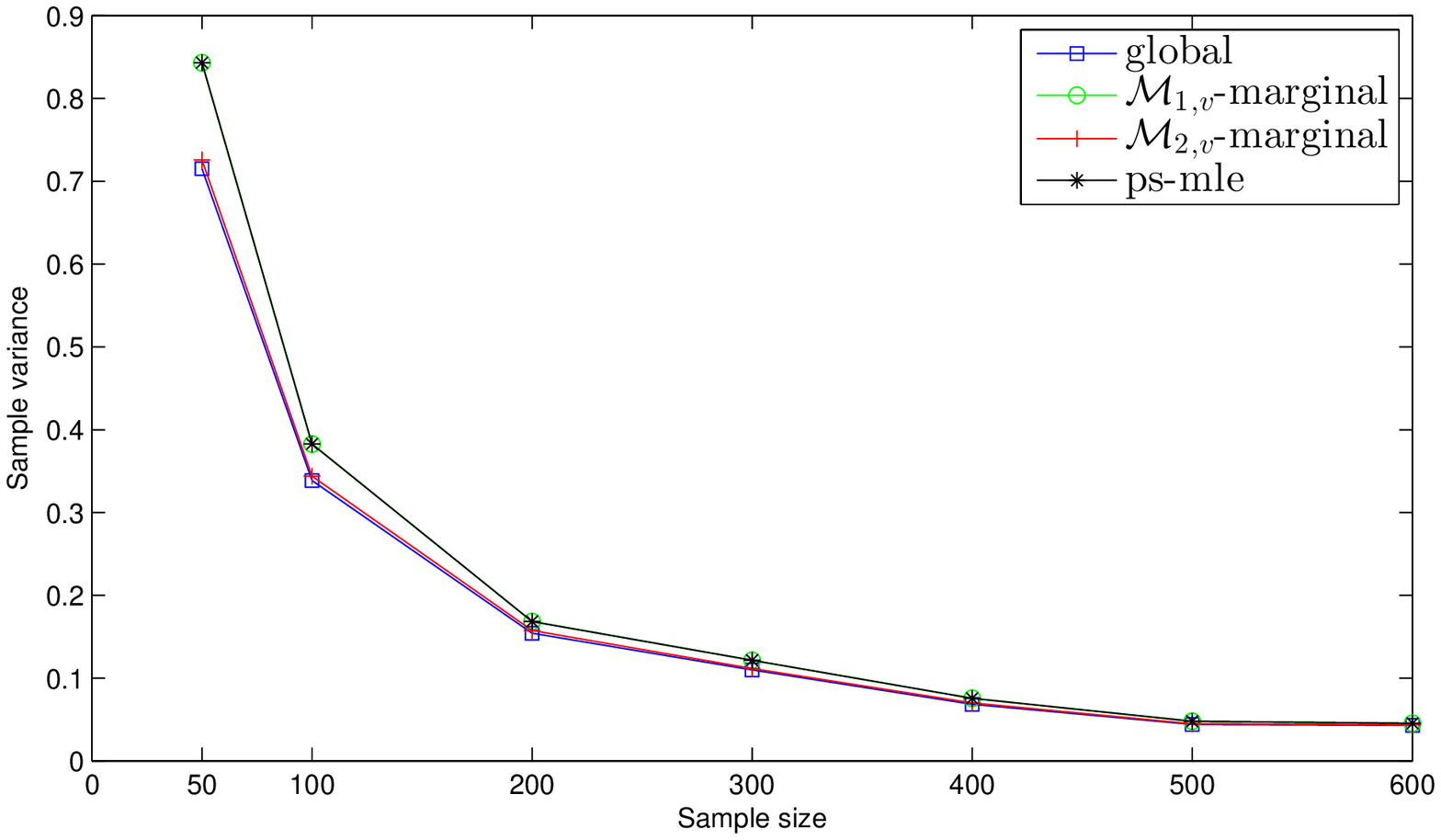}
\caption{Plot of the variance of the relaxed $\M_{i,v},\;i=1,2$- marginal estimates $\hat{\theta}^{\M_{i,v}}_j,\;i=1,2$, the global estimate $\hat{\theta}$ and the pseudo-likelihood estimate $\hat{\theta}^{ps-mle}$ in the four-neighbour $10\times 10$ lattice.}
\label{var10}
\end{figure}

\begin{figure}[htp]
\centering
  \includegraphics[width=15cm]{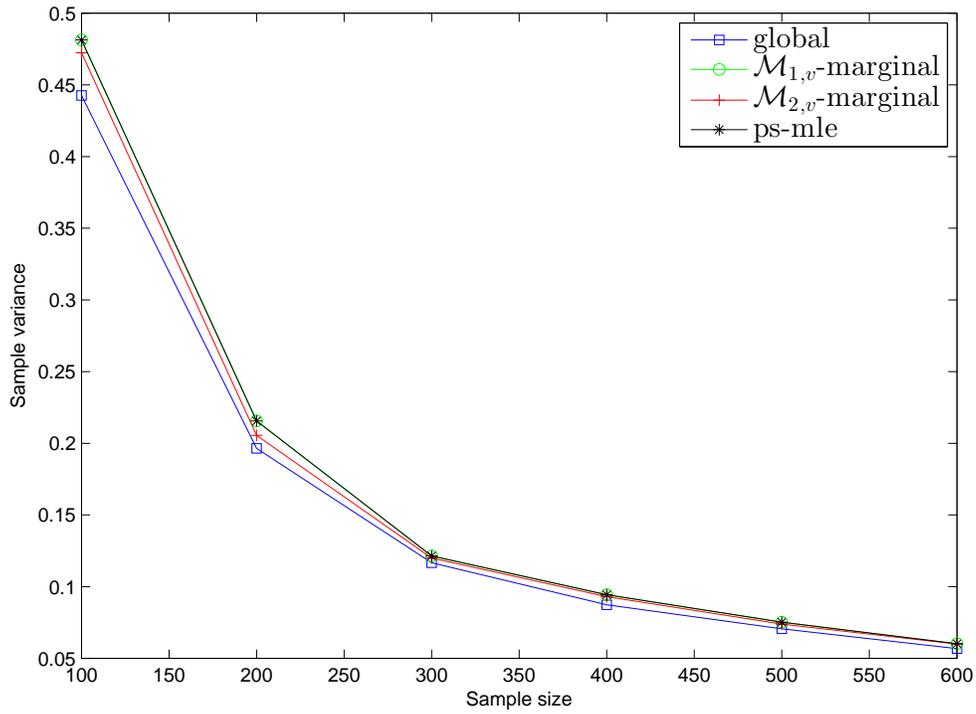}
\caption{Plot of the variance of the relaxed $\M_{i,v},\;i=1,2$- marginal estimates $\hat{\theta}^{\M_{i,v}}_j,\;i=1,2$, the global estimate $\hat{\theta}$ and the pseudo-likelihood estimate $\hat{\theta}^{ps-mle}$ for the graph in Fig. \ref{randomg}.}
\label{varrandomg}
\end{figure}

We now compare the accuracy of the estimates by looking at their sample variance. As predicted by Theorem \ref{variance}, the sample variance of the two-hop relaxed marginal estimate is much smaller than the variance of the one-hop marginal and pseudo-likelihood estimates and  inequalities
\eqref{var} also hold for the sample variances. 
We consider three different graphs, the four-neighbor lattice, $k=4,10$ and a random graph as illustrated in Figure \ref{randomg}. For each graph, we compute the variance of the estimates of a particular $\theta_j, S(j)\subset \M_v, S(j)\not \subset \B_v$ for a vertex $v$ having 11, 13 and 13  neighbors in their two-hop neighborhoods, respectively. For the $k=10$ lattice and the random graph, the global mle is computed with a standard approximate method using belief propagation. The results are given in Figures \ref{var7}, \ref{var10} and \ref{varrandomg} respectively. We see that for $k=4$, the global mle and the two-hop estimate have identical variances. For $k=10$, the variance of the global mle is slightly smaller than that of the two-hop marginal estimate while for the random graph, we see a definite difference between the global mle and the two-hop estimate. Still the variance of the two-hop estimate is always better than that of the one-hop or pseudo-likelihood estimates though the difference gets smaller as the sample size increases. This concurs with the observations of Meng et al. (2013) for the Gaussian case.


\subsection{A small real world example}
\begin{figure}[htp]
\centering
  \includegraphics[width=15cm]{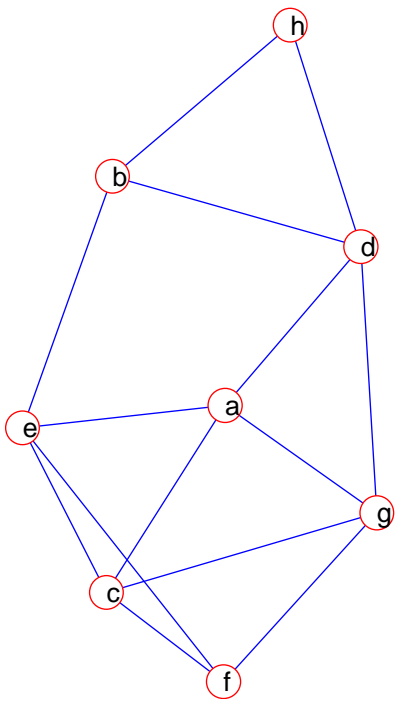}
\caption{The graph underlying the model used for the Rochdale data.}
\label{rochdalegraph}
\end{figure}

In this section, we consider the Rochdale data which has been well studied in Statistics. The reader is referred to Whittaker (1990) for a description of the data. The variables and the graphical model underlying the discrete graphical model are represented in Figure \ref{rochdalegraph}. The purpose of giving this example is simply to illustrate again the accuracy of the two-hop marginal estimate.

The four estimates for the various parameters are computed and we find that the sum of the  absolute value of the differences between the $\M_{1,v}$-, $\M_{2,v}$- , pseudo-likelihood estimates and the global mle are respectively
   $$0.3437,\;\;1.2177\times 10^{-07},\;\;0.2484.$$
   We see that the the pseudo-likelihood estimate is slightly closer to the global mle than the one-hop estimate but that the two-hop estimate is extremely close to the mle.
\vspace{2mm}

\section{Conclusion}
In this paper, we have defined the one-hop and two-hop local marginal estimates for the canonical parameter of a discrete loglinear model Markov with respect to a graph. The one-hop marginal mle is faster to obtain  and just as accurate as the pseudo-likelihood mle for large non-decomposable graphs such as the four-neighbor lattice. Like the latter, it can be improved upon using the methods described in Liu and Ihler (2012) using the local marginal likelihoods rather than the local pseudo-likelihoods. The two-hop marginal estimate is clearly the slowest of the three methods but also the most accurate. Possibly the most important result of this paper is having shown the inequalities between the variances of the one-hop, two-hop and global mle. 
The two-hop estimate is therefore certainly the most accurate with the tightest variance among the local estimates discussed in this paper.

\section{References}

\noindent Boyd, S., Parikh, N., Chu, E., Peleato, B. and Eckstein, J., (2010), Distributed optimization and statistical learning via the alternating direction method of multipliers, {\it Foundations and Trends in Machine Learning}, {\bf 3}, 1-122.
\vspace{2mm}

\noindent Fienberg, S. E. and Rinaldo, A. (2012). Maximum likelihood estimation in log-linear models. {\it Ann. Statist.}, {\bf 40}, 996Ð1023. 
\vspace{2mm}

\noindent  V. Ganapathi, D. Vickrey, J. Duchi, and D. Koller (2008). Constrained Approximate Maximum Entropy Learning. Proceedings of the Twenty-fourth Conference on Uncertainty in AI (UAI). Also Arxiv 1206.3257 (2012)
\vspace{2mm}

\noindent Lauritzen, S.L. (1996),  \textit{Graphical Models}, Oxford Science Publications.
\vspace{2mm}

\noindent Letac, G. and Massam, H., (2012),  Bayes regularization and the geometry of discrete hierarchical loglinear models,  {\it The Annals of Statistics}, {\bf 40}, 861-890.
\vspace{2mm}

\noindent  Liu, Q.  and Ihler, A., (2012), Distributed parameter estimation via pseudo-likelihood, {\it  International Conference on Machine Learning, (ICML)}, June 2012.
\vspace{2mm}

\noindent Meng, Z., Wei, D. Wiesel, A. and Hero, A.O. III, (2013), Distributed learning of Gaussian graphical models via marginal likelihood, {\it Journal of Machine Learning Research}, {\bf 31}, 39-47.
\vspace{2mm}

\noindent Whittaker, J. (1990).  \textit{Graphical Models in Applied Multivariate Statistics}, John Wiley \&
Sons.
\vspace{2mm}

\noindent Wiesel, A. and Hero, A.O. III, (2012), Distributive covariance estimation in Gaussian graphical models, {\it IEEE Transactions on signal processing}, {\bf 60}, 211-220.
\vspace{2mm}

\end{document}